\documentclass[letterpaper]{article} 
\usepackage{aaai25}  
\usepackage{times}  
\usepackage{helvet}  
\usepackage{courier}  
\usepackage[hyphens]{url}  
\usepackage{graphicx} 
\usepackage{natbib}  
\usepackage{caption} 
\usepackage{multirow}
\usepackage{booktabs}
\usepackage{array}
\usepackage{algorithm}
\usepackage{algorithmic}
\usepackage{amsmath}
\usepackage{amssymb}
\usepackage{newfloat}
\usepackage{listings}
\DeclareCaptionStyle{ruled}{labelfont=normalfont,labelsep=colon,strut=off} 
\floatstyle{ruled}
\newfloat{listing}{tb}{lst}{}
\floatname{listing}{Listing}
\title{Alleviating Performance Disparity in Adversarial Spatiotemporal Graph Learning Under Zero-Inflated Distribution}
\author{
    Songran Bai\textsuperscript{\rm 1,\rm 2}, Yuheng Ji\textsuperscript{\rm 1,\rm 2}, Yue Liu\textsuperscript{\rm 3}, Xingwei Zhang\textsuperscript{\rm 1,\rm 2}, Xiaolong Zheng\textsuperscript{\rm 1,\rm 2}\thanks{Corresponding author}, Daniel Dajun Zeng\textsuperscript{\rm 1,\rm 2}\\
}
\affiliations{
    \textsuperscript{\rm 1}Institute of Automation, Chinese Academy of Sciences\\
    \textsuperscript{\rm 2}School of Artificial Intelligence, University of Chinese Academy of Sciences\\
    \textsuperscript{\rm 3}Institute of Data Science \& School of Computing, National University of Singapore\\

    songran.bai@mais.ia.ac.cn\\
    
}

\usepackage{bibentry}

\begin{document}

\maketitle

\begin{abstract}
Spatiotemporal Graph Learning (SGL) under Zero-Inflated Distribution (ZID) is crucial for urban risk management tasks, including crime prediction and traffic accident profiling. However, SGL models are vulnerable to adversarial attacks, compromising their practical utility. While adversarial training (AT) has been widely used to bolster model robustness, our study finds that traditional AT exacerbates performance disparities between majority and minority classes under ZID, potentially leading to irreparable losses due to underreporting critical risk events. In this paper, we first demonstrate the smaller top-k gradients and lower separability of minority class are key factors contributing to this disparity. To address these issues, we propose \underline{MinGRE}, a framework for \underline{Min}ority Class \underline{G}radients and \underline{R}epresentations \underline{E}nhancement. MinGRE employs a multi-dimensional attention mechanism to reweight spatiotemporal gradients, minimizing the gradient distribution discrepancies across classes. Additionally, we introduce an uncertainty-guided contrastive loss to improve the inter-class separability and intra-class compactness of minority representations with higher uncertainty. Extensive experiments demonstrate that the MinGRE framework not only significantly reduces the performance disparity across classes but also achieves enhanced robustness compared to existing baselines. These findings underscore the potential of our method in fostering the development of more equitable and robust models.
\end{abstract}

%

\section{Introduction}

Spatiotemporal Graph Neural Networks (STGNNs) have emerged as a vital component in modeling complex spatiotemporal dependencies within Spatiotemporal Graph Learning (SGL) under the Zero-Inflation Distribution (ZID) \cite{LiuTKDE, LiuTKDD, ZhaoTITS, TriratTITS, CIKMSTGNN}. The datasets that conforms to such distribution consist of a majority of zero observations and a minority of non-zero observations \cite{ZIDKDD, ZIDWWW, ZeroInflated2006, ZeroInflated2021}. Effectively addressing ZID is pivotal for discerning sparse event patterns in urban crime analysis, traffic accident forecasting, and demand prediction \cite{ZhuangKDD, WangTITS, AAAILiu2024, KDDMOE}.

Nevertheless, recent studies have identified vulnerabilities within STGNNs, where adversaries could induce incorrect traffic predictions by slightly perturbing historical data \cite{ZhuIOT, TNDS, LiICASSP}. Consequently, Adversarial Training (AT) has been introduced to bolster the robustness of these models \cite{RDAT}. This process generally encompasses three key stages: the selection of salient victim nodes, the generation of Adversarial Examples (AEs), and iterative optimization \cite{TNDS, RDAT}. However, the effectiveness of such spatiotemporal adversarial training has been evaluated primarily on dense datasets with normal distributions. Its effectiveness on sparse, zero-inflated datasets remains a significant and worthy area of exploration.

\begin{figure}[!htbp]
  \centering
  \includegraphics[width=\columnwidth]{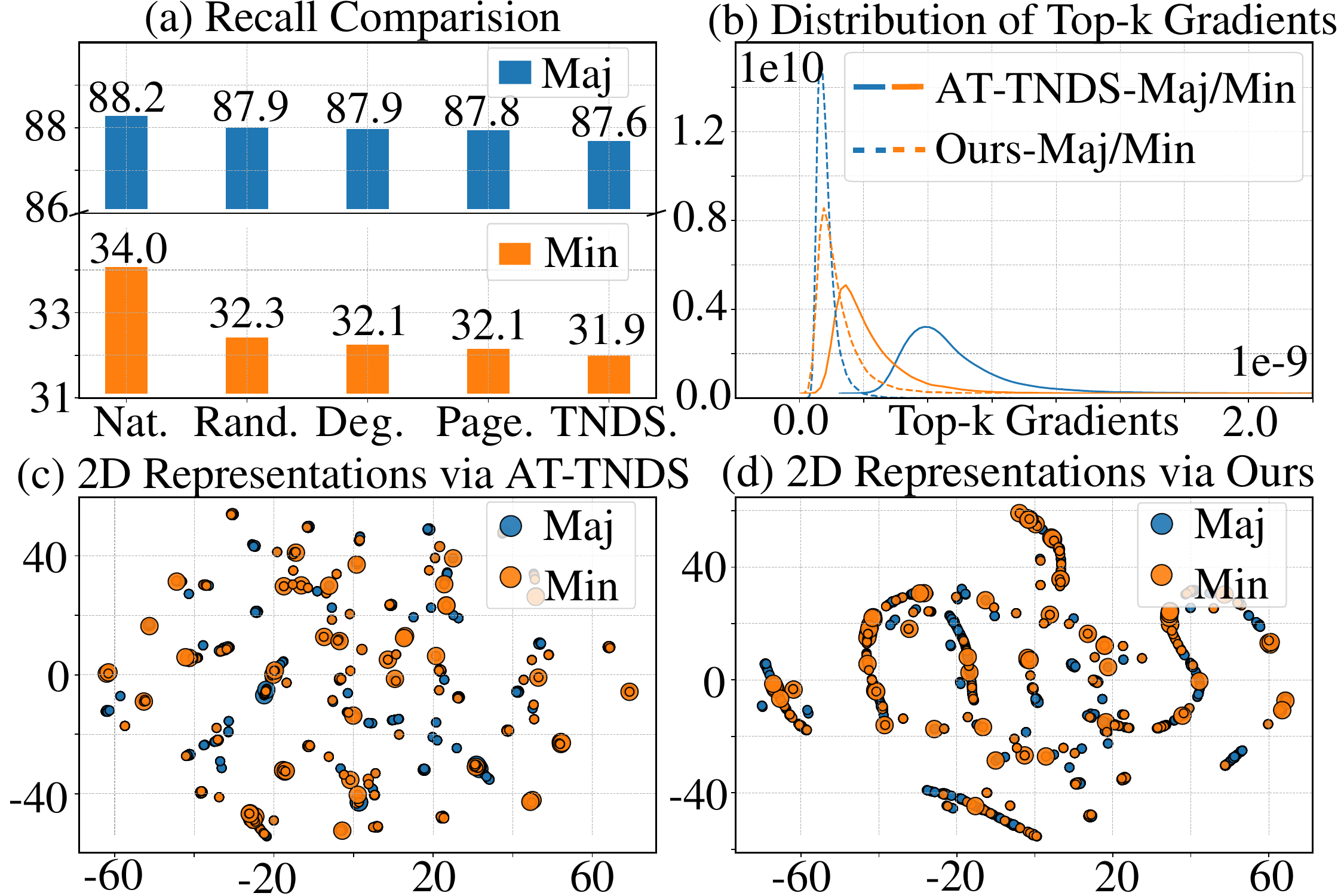}
  \caption{Impact analysis of spatiotemporal adversarial training on the ZID dataset NYC. (a) compares recall metrics between natural training and adversarial training approaches. (b) displays the distribution of top-K gradients for both majority and minority classes throughout the adversarial training. Panels (c) and (d) present two-dimensional projections of the learned features for majority and minority classes via AT-TNDS and our proposed method, respectively.}
  \label{fig:emprical}
\end{figure}
To this end, we initially investigate the performance of existing spatiotemporal adversarial training methods in ZID scenarios, with a particular focus on the prediction performance and robustness of non-zero observations representing minority class, as this is crucial for addressing serious safety concerns such as incident underreporting \cite{Underreporting} in real-world applications. Our empirical analysis of zero-inflated datasets revealed three findings as follows. 1) Conventional spatiotemporal adversarial training approaches tend to exacerbate the performance disparity between majority and minority classes, as illustrated in Figure \ref{fig:emprical}(a), due to a more significant degradation of the minority class. 2) Our study further reveals that the top-k gradients of the minority class are generally weaker, leading to a dominance of majority class adversarial examples in training (see Figure \ref{fig:emprical}(b)). 3) Furthermore, as illustrated in Figure \ref{fig:emprical}(c), the separability of the minority representations deteriorates following adversarial training. Moreover, samples with high uncertainty exhibit greater prediction errors, as indicated by the size of the points in the figure. Thus, we posit that the smaller top-k gradients and lower separability of the minority class are two potential underlying causes of performance disparity.

To address current challenges, we propose the \underline{Min}ority Class \underline{G}radients and \underline{R}epresentations \underline{E}nhancement (MinGRE) framework. Our approach begins with a victim node selection strategy during adversarial training, crucial for generating fairer and more effective perturbations across classes. This strategy employs a cross-segment spatiotemporal encoder to capture complex inter-segment, intra-segment, and spatial dependencies. Additionally, we introduce a multi-dimensional attention-based gradient reweighting technique that adaptively adjusts spatiotemporal gradients throughout the training, reducing bias towards the majority class. Furthermore, inspired by Zha et al.'s work on maintaining continuity in representation space for regression tasks \cite{RegNIPS}, we incorporate an uncertainty-guided contrastive learning loss. This loss function maximizes feature dissimilarity between classes, particularly in regions with high predictive uncertainty.

The main contributions of this paper are as follows:
\begin{itemize}
    \item We analyze the adversarial robustness of SGL models under zero-inflated settings, identifying significant issues in performance disparity.
    \item We introduce a multi-dimensional attention-based gradient reweighting method to improve the selection of victim nodes in spatiotemporal adversarial training.
    \item We employ an uncertainty-guided contrastive loss to focus on representation learning in regression tasks, thereby reducing inter-class similarity and enhancing intra-class cohesion.  
    \item Extensive experiments across various target models, attack methods, and datasets confirm the effectiveness of our proposed framework on both the robustness and disparity metrics.
\end{itemize}

\section{Related Work}
\subsection{Spatiotemporal Graph Learning Under ZID}
Spatiotemporal graph learning has garnered substantial interest, particularly in domains with sparse or zero-inflated data \cite{TITS2024}. Models like GMAT-DU \cite{ZhaoTITS} and RiskSeq \cite{ZhouTDE} underscore the value of granular spatiotemporal data in data-scarce environments \cite{CIKMAccident}. The recent trend of employing graph neural networks (GNNs) with dynamic and multi-view approaches, exemplified by MADGCN \cite{WuDyn} and MG-TAR \cite{TriratTITS}, demonstrates the synergy between spatiotemporal dynamics and attention mechanisms to improve prediction accuracy. Furthermore, the integration of uncertainty quantification in STGNNs \cite{UncArxiv, UncWSDM, SAUC, Uncertainty2024} underscores the necessity for robust models capable of handling sparse data and providing dependable predictions.

\subsection{Adversarial Robustness of Spatiotemporal Graph Learning}
Adversarial attacks are crucial for assessing model robustness \cite{Zhang, ji2024advlora}, especially in spatiotemporal contexts \cite{TNNLSLiu}. Designing such attacks involves dynamically selecting victim nodes and generating time-dependent perturbations while ensuring the attacks remain imperceptible. \cite{ZhuIOT} proposed a query-based black-box attack using SPSA \cite{ICMLSASP} for gradient estimation and a knapsack greedy algorithm for node selection. \cite{TNDS} introduced STPGD, an iterative method suitable for both white-box and gray-box scenarios. ADVERSPARSE \cite{LiICASSP}, on the other hand, targets graph structures by sparsifying them to disrupt spatial dependencies and increase prediction errors. Adversarial training has also shown promise in enhancing robustness \cite{CIKMAT}, with AT-TNDS integrating spatiotemporal perturbations into the training process \cite{TNDS}. \cite{RDAT} leveraged reinforcement learning for dynamic node selection, alongside knowledge distillation to stabilize the policy network. \cite{ICMLzhang} further strengthened spatiotemporal representations using contrastive loss within a self-supervised learning framework. However, current research primarily addresses dense, continuous data, often overlooking the discrete and sparse nature of critical spatiotemporal data \cite{SUSTeR, CIKM2021}.

\subsection{Adversarial Training in Imbalanced Settings}
Recent studies have underscored the critical impact of data imbalance on the effectiveness of adversarial training \cite{CSUR, CVPRATLT, IEEETIT}. In such conditions, adversarial training can amplify the imbalance, thereby diminishing the model’s performance on underrepresented classes \cite{ICDMWang}. To address these challenges, approaches such as scale-invariant classifiers and two-stage rebalancing frameworks have been proposed \cite{CVPRWu}. Furthermore, meta-learning-based sample-aware re-weighting has demonstrated potential in enhancing adversarial robustness within imbalanced datasets \cite{AAAIHou}. These methods aim to balance class representation during training, with strategies like margin engineering and re-weighting showing promise in enhancing adversarial robustness under imbalanced settings \cite{MLText}.

\section{Preliminaries}
\subsection{Spatiotemporal Prediction Under ZID}
Let $\mathcal{G}_{t}^{k} = (\mathcal{V},\mathcal{E}^{k},\mathcal{A}^{k},\mathcal{X}_{t})$ denote multi-view undirected graphs at step $t$, where $\mathcal{V}$ is the set of $N$ nodes that is time-invariant. $\mathcal{E}^{k}$ denotes the edge set of $k^{th}$ view and $\mathcal{A}^{k}$ denotes the adjacency matrix of $k^{th}$ view. Then $\mathcal{X}_{t} \in \mathbb{R}^{N * D}$ denotes the D-dimensional node features at time $t$. The prediction model aims to estimate future node states $\mathcal{Y}_{t+1:t+\Delta}$ as follows:
\begin{equation}
    \mathcal{\hat{Y}}_{t+1:t+\Delta}=f_{\theta}\left(\mathcal{X}_{t-\mathcal{T}+1:t}, \mathcal{A}\right)
\end{equation}
Here, $\mathcal{Y}_{t+1:t+\Delta}$ exhibits the characteristic of zero-inflation distribution, which means that non-zero labels are sparsely distributed in both temporal and spatial dimensions. For simplicity, we will use $\mathcal{X}_{t}^{\mathcal{T}} \in \mathbb{R}^{\mathcal{T} * N * D}$ to represent node features from time $t-\mathcal{T}+1$ to time $t$ in the following content. And we use $\mathcal{Y}_{t}^{\Delta}$, $\hat{\mathcal{Y}}_{t}^{\Delta} \in \mathbb{R}^{\Delta * N}$ to represent the real and predicted node states from time $t+1$ to time $t+\Delta$. The widely used weighted RMSE loss function \cite{TKDERMSE} can be defined as:
\begin{equation}
    \mathcal{L}\left(\mathcal{\hat{Y}}_{t}^{\Delta},\mathcal{Y}_{t}^{\Delta}\right)=\frac{1}{\Delta * N}\sum_{\substack{\delta,n}}w_{t}^{\left(\delta,n\right)}\left(y_{t}^{\left(\delta,n\right)}-\hat{y}_{t}^{\left(\delta,n\right)}\right)^{2}
\end{equation}
where $y_{t}^{\left(\delta,n\right)}$, $\hat{y}_{t}^{\left(\delta,n\right)}$ and $w_{t}^{\left(\delta,n\right)}$ represent the real state, the predicted state and the loss weight of node $\mathcal{V}_{n}$ at time $t+\delta$, respectively. 
\subsection{Spatiotemporal Adversarial Attack}
The objective of spatiotemporal graph adversarial attacks is to maximize prediction errors by perturbing the historical attributes of a minimal subset of node features. The optimal AEs can be defined as \cite{TNDS}:
\begin{equation}
     \operatorname*{argmax}_{\substack{\left(\mathcal{X}_{t}^{\mathcal{T}}\right)^{\prime} \in \mathcal{B}\left(\mathcal{X}_{t}^{\mathcal{T}} \right)}} \sum_{t\in T_{test}}\mathcal{L}\left(f_{\theta^{*}}\left(\cdot\right),\mathcal{Y}_{t}^{\Delta}\right) 
\end{equation}
\begin{equation}
     s.t. \ \ \left\|\left(\left(\mathcal{X}_{t}^{\mathcal{T}}\right)^{\prime} -\mathcal{X}_{t}^{\mathcal{T}}\right)\circ\mathcal{P}_{t}\right\|_{p} \leq \epsilon, \ \left\|\mathcal{P}_{t}\right\|_{0} \leq \eta * N
\end{equation}
where $\epsilon$ and $\eta$ denote the attack budget and the proportion of nodes being attacked, respectively. And $\mathcal{P}_{t} \in \mathbb{R}^{1 * N * 1}$ represents a three-dimensional matrix containing only 0 and 1. If the elements $\mathcal{P}_{t}\left(:,i,:\right)$ are 1, it indicates that the node $\mathcal{V}_{i}$ will be attacked. And $\mathcal{B}\left(\mathcal{X}_{t}^{\mathcal{T}}\right)=\left\{\mathcal{X}_{t}^{\mathcal{T}}+\Phi_{t}^{\mathcal{T}}\circ\mathcal{P}_{t} \mid \|\Phi_{t}^{\mathcal{T}}\circ\mathcal{P}_{t}\|_{p} \leq \epsilon \right\}$ represents the allowed perturbation set. To solve the above optimization problem, \cite{TNDS} firstly calculate the gradient-based time-dependent non-negative node saliency within a batch:
\begin{equation}
    \mathcal{S}_{T_{batch}} = \|\text{Relu}\left(\frac{1}{B}\sum_{t \in T_{batch}}\nabla\mathcal{L}\left(\cdot\right)\right)\|_{2}
\end{equation}
then the victim nodes can be represented by $\mathcal{P}_{t}\left(:,i,:\right)=\boldsymbol{1}_{\mathcal{V}_{i} \in \text{top}_{k}\left(\mathcal{S}_{\mathcal{T}_{batch}}\right)}$, where $\mathcal{P}_{t}\left(:,i,:\right)$ will be 1 if $\mathcal{V}_{i}$ is the $top-k$ salient node within a batch $\mathcal{T}_{batch}$. Based on the victim nodes, the iteration process of Spatiotemporal Projected Gradient Desent (STPGD) can be defined as:
\begin{equation}
    \left(\mathcal{X}_{t}^{\mathcal{T}}\right)^{\prime\left(i\right)} = clip_{\epsilon}\left(\left(\mathcal{X}_{t}^{\mathcal{T}}\right)^{\prime\left(i-1\right)}+\alpha \text{sign}\left( \nabla \mathcal{L}\left(\cdot \right)\circ\mathcal{P}_{t}\right)\right)
\end{equation}
where $\text{clip}_{\epsilon}\left(\cdot\right)$ is the operation to bound the perturbation in a $\epsilon$ ball. And $\left(\mathcal{X}_{t}^{\mathcal{T}}\right)^{\prime\left(i\right)}$ represents the adversarial features of $i^{th}$ iteration.
\subsection{Spatiotemporal Adversarial Training}
Adversarial training in the context of spatiotemporal graph learning can also be regarded as a min-max optimization process, which enhances the robustness of the model against adversarial attacks. This can be formulated as:
\begin{equation}
    \min_{\theta}\max_{\left(\mathcal{X}_{t}^{\mathcal{T}}\right)^{\prime} \in \mathcal{B}\left(\mathcal{X}_{t}^{\mathcal{T}}\right)} \sum_{t \in T_{train}}\mathcal{L}\left(f_{\theta}\left(\cdot\right),\mathcal{Y}_{t}^{\Delta}\right)
\end{equation}
Since the above problem is most likely a non-convex bi-level optimization problem, many studies approximate it by alternating first-order optimization, that is, training $f_{\theta}$ on the adversarial perturbed spatiotemporal graph in each iteration.

\section{Methodology}
This section delineates the MinGRE framework through two key components: the Adversarial Examples Generation Module and the Uncertainty-guided Contrastive Loss Module, as illustrated in Figure \ref{fig:frame}. The implementation of our proposed method is presented in the Appendix.
\begin{figure*}[t] %
  \centering
  \includegraphics[width=\textwidth]{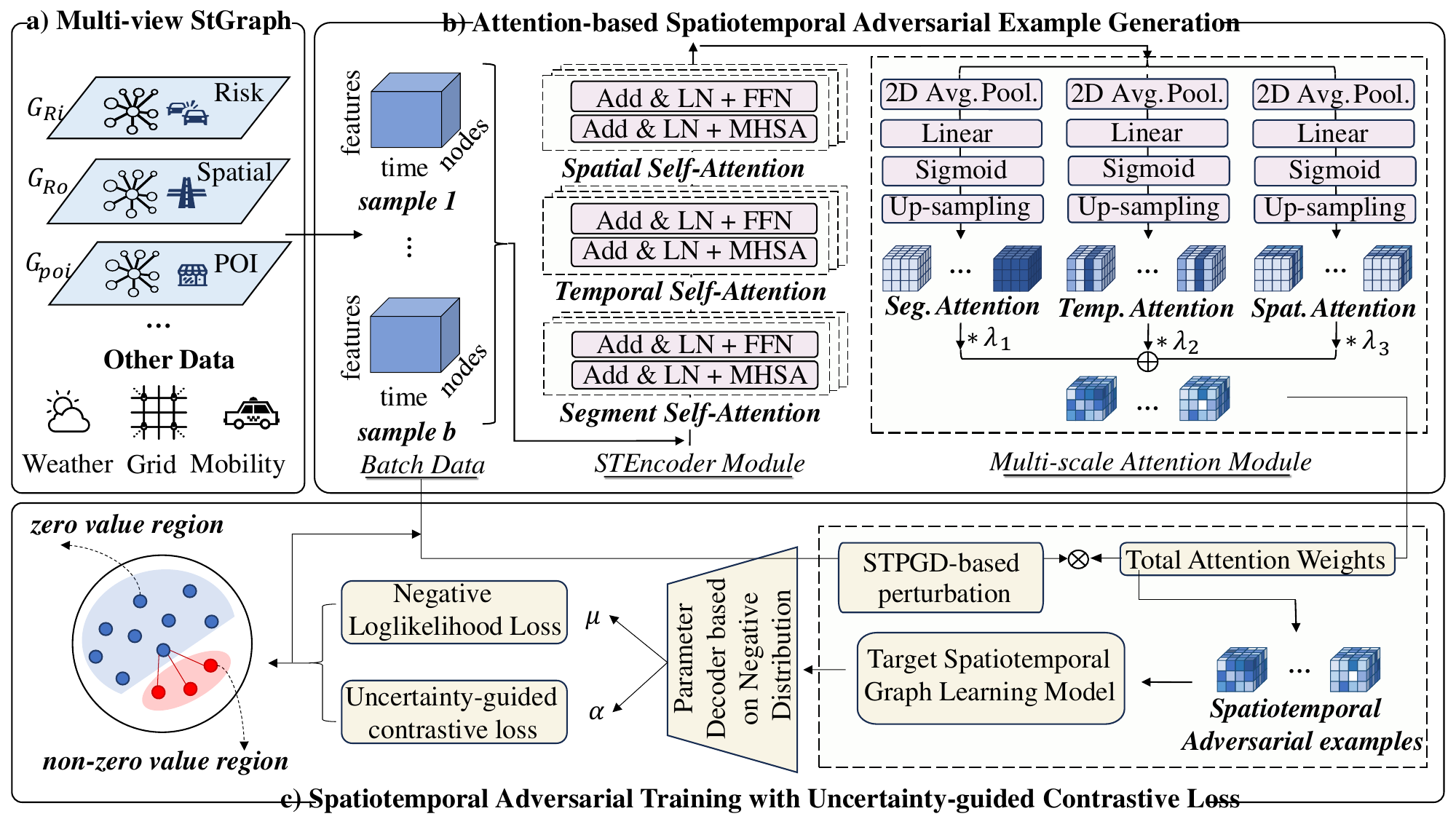} 
  \caption{The overall framework of our proposed MinGRE.}
  \label{fig:frame}
\end{figure*}

\subsection{Adversarial Examples Generation Module}
The primary challenge in generating adversarial samples is effectively reweighting gradients to ensure a more balanced selection of victim nodes. For instance, considering the weighted RMSE Loss, we can simplify the expression for the gradient $\nabla\mathcal{L}\left(\cdot\right)$ of a sample $i$ using the chain rule, as follows:
\begin{equation}
    \nabla\mathcal{L}\left(\cdot\right)=\frac{\partial\mathcal{L}}{\partial \hat{y}_{i}}\cdot\frac{\partial \hat{y}_{i}}{x_{i}}=w_{i}\cdot\frac{\partial \hat{y}_{i}}{x_{i}}=w_{i}\cdot \mathcal{G}_{i} \\
    \label{grad}
\end{equation}
It can be observed from the above equation that the predefined weight \(w_{i}\) and the variable \(\mathcal{G}_{i}\) determine the final magnitude of the gradients. Notably, \(w_{i}\) is set based on expert knowledge, and \(\mathcal{G}_{i}\) assumes that gradient flow across different temporal dimensions of \(x_{i}\) holds uniform importance \cite{NEURIPS2021}. However, node selection strategies based on these assumptions are unsuitable for ZID scenarios, as they can result in biased gradient distributions between majority and minority classes. To address this issue, we propose a multi-dimensional gradient reweighting strategy that employs segment and spatial attention to focus on samples within specific segments and nodes. Additionally, we introduce temporal attention mechanisms to differentiate the importance of gradient flows from various temporal dimensions of the input feature \(\mathcal{X}\). This approach is implemented through three key components: a Cross-Segment Spatiotemporal Encoder, Gradient Reweighting-based Adversarial Example Generation, and an Optimization Objective.

\subsubsection{Cross-Segment Spatiotemporal Encoder}
Building on the work of \cite{ABDNIPS}, we have incorporated the Attention Between Datapoints (ABD) mechanism to capture pairwise interactions across different segments within a batch.
Consider a batch of spatiotemporal segments denoted as $\mathcal{X} = \left\{\mathcal{X}_t^{\mathcal{T}} \in \mathbb{R}^{\mathcal{T} * N * D} \mid t = t_{1}, \dots, t_{B}\right\}$. The ABD layer processes these samples as follows:
\begin{equation}
\mathcal{O}_{sg}\left(\mathcal{X}\right) = \text{LN}\left(\mathcal{R}\left(\mathcal{X}_{sg}\right) + \text{FFN}\left(\mathcal{R}\left(\mathcal{X}_{sg}\right)\right)\right), \\
\label{ABD}
\end{equation}
where $\mathcal{X}_{sg} = \pi_{\sigma\left(sg\right)}\left(\mathcal{X}\right)$ reshapes the input tensor $\mathcal{X}$ to conform to the dimensions $\left(\mathcal{T},N,B,D\right)$. The function $\mathcal{R}\left(\mathcal{X}_{sg}\right)$, defined by
\begin{equation}
\mathcal{R}(\mathcal{X}_{sg}) = \text{LN}\left(\mathcal{M}\left(\mathcal{X}_{sg}\right) + \mathcal{X}_{sg}\right), \\
\label{Res}
\end{equation}
represents the residual output of the ABD module. Here, $\mathcal{M}\left(\mathcal{X}_{sg}\right)$ computes the output from the multi-head self-attention mechanism, which is expressed as:
\begin{equation}
\mathcal{M}\left(\mathcal{X}_{sg}\right) = \text{concat}\left(\mathcal{\mathcal{M}}_{sg}^1, \dots, \mathcal{M}_{sg}^k\right) \mathcal{W}_{sg}^\mathcal{M}, \\
\label{MHSAtt}
\end{equation}
where each $\mathcal{M}_{sg}^j$ is obtained by \\
\label{Softmax}
\begin{equation}
\mathcal{M}_{sg}^j = \text{softmax}\left(\frac{\mathcal{Q}_{sg}^j \left(\mathcal{K}_{sg}^j\right)^T}{\sqrt{d}}\right) \mathcal{V}_{sg}^j,
\end{equation}
and embedding matrices $\left(\mathcal{Q}_{sg}^j, \mathcal{K}_{sg}^j, \mathcal{V}_{sg}^j\right)$ are computed as $\left(\mathcal{X}_{sg} \mathcal{W}_{sg}^{\mathcal{Q}_j}, \mathcal{X}_{sg} \mathcal{W}_{sg}^{\mathcal{K}_j}, \mathcal{X}_{sg} \mathcal{W}_{sg}^{\mathcal{V}_j}\right).$
This formulation effectively models complex interactions between the segments.

Similarly, in order to sequentially encode the temporal and spatial dependencies \cite{STAEformer}, we connect the temporal self-attention and spatial self-attention mechanisms in series after the ABD module. The computation process is the same as equation (\ref{ABD}), and the final output of the encoder is:
\begin{equation}
    \mathcal{O}\left(\mathcal{X}\right) = \mathcal{O}_{sp}\left(\pi_{\sigma\left(sp\right)}\left(\mathcal{O}_{te}\left(\pi_{\sigma\left(te\right)}\left(\mathcal{O}_{sg}\left(\mathcal{X}\right)\right)\right)\right)\right),
\end{equation}
where $\pi_{\sigma\left(te\right)}\left(\cdot\right)$ reshapes the input tensor to conform to the dimensions $\left(B,N,\mathcal{T},D_{h}\right)$ and $\pi_{\sigma\left(sp\right)}\left(\cdot\right)$ reshapes the input tensor to conform to the dimensions $\left(B,\mathcal{T},N,D_{h}\right)$.

\subsubsection{Gradient Reweighting-Based Adversarial Example Generation}
We aim to increase the selection probability of non-zero regions in adversarial sample generation. A potential solution is to re-weight the spatiotemporal gradients $grad=\left\{\nabla \mathcal{L}\left(\cdot \right) \in \mathbb{R}^{\mathcal{T} \times N \times D} \mid t=t_{1}, \dots, t_{B}\right\}$ during the iterative process, skewing the gradient distribution towards non-zero regions. This ensures the top-k node selection strategy targets nodes with more non-zero observations. We propose a learning-based re-weighting method using a multi-dimensional attention mechanism, integrating segment, temporal, and spatial attention matrices. The segment attention weight matrix $Att_{sa}$, inspired by channel attention mechanisms \cite{ChannelCVPR}, is computed as follows:
  \begin{equation}
     Att_{sg}=\mathcal{C}\left(\sigma\left(g_{3}^{sg}\left(g_{2}^{sg}\left(g_{1}^{sg}\left(\text{Pool}_{\mathcal{T},N}\left(\mathcal{O}\left(\mathcal{X}\right)\right)\right)\right)\right)\right)\right) \\
     \label{Att_sa}
 \end{equation}
 Here we first perform two-dimensional pooling compression $\text{Pool}_{\mathcal{T},N}$ in the temporal and spatial dimensions to obtain a matrix of shape $\left(B,1,1,D_{h}\right)$. Subsequently, we derive a weight matrix of shape $\left(B,1,1,1\right)$ based on a three-layer perceptron $\left(g_{1}^{sg}\left(\cdot\right), g_{2}^{sg}\left(\cdot\right), g_{3}^{sg}\left(\cdot\right)\right)$ and a Sigmoid layer $\sigma\left(\cdot\right)$, which represents the significance of different segments. Ultimately, the elements of this weight matrix are replicated and expanded by $\mathcal{C}\left(\cdot\right)$ to form a weight matrix of shape $\left(B,\mathcal{T},N,D\right)$, reflecting the weight distribution across the original spatiotemporal gradients. Temporal and spatial attention are also similar to (\ref{Att_sa}). So the final gradients after reweighting can be denoted as:
 \begin{equation}
     \hat{grad} = Att_{1} \circ grad \circ Att_{te}
 \end{equation}
 where $Att_{1} = Att_{sg} + Att_{sp}$ can be used to correct $w_{i}$ in (\ref{grad}), and $Att_{te}$ can be used to reweight $\mathcal{G}_{i}$ in (\ref{grad}). Thus, the new attention-guided spatiotemporal graph adversarial sample generation process can be described as follows: 
\begin{equation}
    \mathcal{P}\left(:,:,i,:\right)=\boldsymbol{1}_{\mathcal{V}_{i} \in \text{top}_{k}\left(\\|\text{Relu}\left(\hat{grad}\right)\|_{2}\right)}
\end{equation}
\begin{equation}
    \mathcal{X}^{\prime\left(i\right)} = \text{clip}_{\epsilon}\left(\mathcal{X}^{\prime\left(i-1\right)}+\alpha \text{sign}\left(\hat{grad} \circ \mathcal{P}\right)\right)
\end{equation}
 \subsubsection{Optimization Objective}
In the context of the performance disparity problem studied in this paper, we hope to strengthen the gradient of the minority samples, so we designed a specific optimization loss to guide the reweighting network. Given the coupled nature of adversarial attacks and the optimization of the reweighting network, we adopt a two-stage iterative strategy for learning. In the first stage, the optimization objective of an adversarial attack is:
 \begin{equation}
     \operatorname*{argmax}_{\substack{\left(\mathcal{X}_{t}^{\mathcal{T}}\right)^{\prime}_{\psi^{*}} \in \mathcal{B}\left(\mathcal{X}_{t}^{\mathcal{T}} \right)}} \sum_{t \in T_{train}}\mathcal{L}\left(f_{\theta^{*}}\left(\left(\mathcal{X}_{t}^{\mathcal{T}}\right)^{\prime}_{\psi^{*}}\right),\mathcal{Y}_{t}^{\Delta}\right)
 \end{equation}
 In the second stage, the optimization objective of reweighting the network is: 
\begin{equation}
    \begin{split}
        \operatorname*{argmin}_{\substack{\psi}} & \sum_{t \in T_{train}} \lambda_{1}\mathcal{L}\left(f_{\theta^{*}}\left(\left(\mathcal{X}_{t}^{\mathcal{T}}\right)^{*}_{\psi}\right),\mathcal{Y}_{t}^{\Delta}\right) \\
        & + \lambda_{2} MAE\left(\left(\hat{grad}_{t}^{\mathcal{T}}\right)_{+},\left(\hat{grad}_{t}^{\mathcal{T}}\right)_{-}\right) \\
        & + \lambda_{3}\|\left(Att_{1t}^{\mathcal{T}}\right)_{+}^{\prime} \|_{2} + \lambda_{4}\|\left(Att_{1t}^{\mathcal{T}}\right)_{-}\|_{2}
    \end{split}
\end{equation}
 where $\left(\hat{grad}_{t}^{\mathcal{T}}\right)_{+}$ and $\left(\hat{grad}_{t}^{\mathcal{T}}\right)_{-}$ respectively represent the spatiotemporal gradients of minority class and majority class. Similarly, $\left(Att_{1t}^{\mathcal{T}}\right)_{+}$ and $\left(Att_{1t}^{\mathcal{T}}\right)_{-}$ respectively represent the weight matrices of majority class and minority class. 
 
\subsection{Uncertainty-Guided Adversarial Contrastive Loss}
Previous studies show that feature separability helps mitigate performance degradation in minority classes during adversarial training in imbalanced classification tasks \cite{ICDMWang}. In regression tasks, \cite{RegNIPS} highlighted the significance of continuous embeddings consistent with labels for enhancing model robustness and generalization. Moreover, mining hard negative and hard positive samples can effectively enhance the model's discriminative ability for these samples \cite{AAAILiu}. Building on this, we introduce an uncertainty-guided supervised contrastive learning approach. Given the abundance of zero-value regions, we prioritize hard-to-distinguish examples using uncertainty quantification based on parameter decoding \cite{NIPSVAE}. For the zero-inflated spatiotemporal data, the negative binomial distribution \cite{NB,ZhuangKDD} is a more appropriate fit than the Gaussian assumption implied by RMSE, with its probability mass function defined as:
\begin{equation}
    \mathcal{P}_{NB}\left(x_{k};n,p\right)=\binom{x_{k}+n-1}{n-1}\left(1-p\right)^{x_{k}}p^{n} \\ \label{NB1}
\end{equation}
where $x_{k}$ and $n=\frac{\mu\alpha}{1-\alpha}$ are the number of failures and successes respectively, and $p=\frac{1}{1+\mu\alpha}$ is the probability of a single success. 

The parameter decoding process based on the negative binomial distribution is as follows:
\begin{equation}
    \left(\hat{\mathcal{\mu}}_{t}^{\Delta}, \hat{\mathcal{\alpha}}_{t}^{\Delta}\right)=f_{decoder}\left(h_{target}\left(\mathcal{X}_{t}^{\mathcal{T}},\mathcal{A}\right)\right)
\end{equation}
where $h_{target}\left(\mathcal{X}_{t}^{\mathcal{T}},\mathcal{A}\right)=\hat{\mathcal{H}}_{t}^{\mathcal{T}}$ is the hidden feature embedding calculated by the target model before the output layer. And $\hat{\mu}_{t}^{\Delta}$ is the mean parameter of the distribution predicted by the decoder network, and $\hat{\alpha}_{t}^{\Delta}$ is the predicted dispersion parameter. We use the variance parameter $\hat{\alpha}_{t}^{\Delta}$ predicted by the decoder as an indicator of the difficulty of the region, and combine it with the supervised contrastive loss \cite{NEURIPS2020CL} as a weight value. The final form of the adversarial training loss used in this paper is:
\begin{equation}
\begin{split}
    \mathcal{L}_{adv} & =\beta_{1}\sum_{t \in T_{train}}\mathcal{L}_{nb}\left(\hat{\mu}_{t}^{\Delta}, \hat{\alpha}_{t}^{\Delta}, \mathcal{Y}_{t}^{\Delta}\right) \\
    & + \beta_{2}\sum_{t \in T_{train}}u_{t}\mathcal{L}_{scl}\left(\hat{\mathcal{H}}_{t}^{\mathcal{T }}\right)
\end{split}
\end{equation}
where $\mathcal{L}_{nb}$ represents the negative log-likelihood loss function based on the negative binomial distribution. And $u_{t}=\frac{2}{1+e^{-\hat{\alpha}_{t}^{\Delta}/\gamma}}-1$ represents the normalized weights based on the uncertainty represented by the predicted variance. And $\mathcal{L}_{scl}\left(\hat{\mathcal{H}}_{t}^{\mathcal{T}}\right)$ represents the supervised contrastive learning loss function based on the feature embedding \cite{CVPRContras}.
\begin{table*}[t]
\centering
\setlength{\tabcolsep}{4pt} 
\renewcommand{\arraystretch}{0.9} 
\resizebox{\textwidth}{!}{
\begin{tabular}{c c c c c c c c c c}
\toprule
\multirow{2}{*}{\centering\arraybackslash\textbf{Dataset}} & \multicolumn{1}{c}{\textbf{Attacks}} & \multicolumn{2}{c}{\textbf{Clean}} & \multicolumn{2}{c}{\textbf{STPGD-TNDS}} & \multicolumn{2}{c}{\textbf{Clean}} & \multicolumn{2}{c}{\textbf{STPGD-TNDS}} \\ 
\cmidrule(lr){2-10}
  & \multicolumn{1}{c}{\textbf{Metrics}} & \textbf{Rec-maj} & \textbf{Rec-min} & \textbf{Rec-maj} & \textbf{Rec-min} & \textbf{MAP-maj} & \textbf{MAP-min} & \textbf{MAP-maj} & \textbf{MAP-min} \\ 
\midrule
\multirow{5}{*}{\centering\arraybackslash\textbf{NYC}} & NT-WRMSE &\underline{88.182}&\underline{33.956}&87.012&27.416&\underline{0.7847}&\underline{0.1869}&0.7580&0.1467\\ 
& AT-Random &87.888&32.308&87.543&30.381&0.7808&0.1817&0.7642&0.1591\\ 
& AT-Degree&87.857&32.138&87.602&30.710&0.7801&0.1824&0.7683&\underline{0.1628}\\ 
& AT-TNDS &87.586&31.893&\underline{87.856}&\underline{30.974}&0.7813&0.1782&\underline{0.7701}&0.1458 \\ 
& \centering\arraybackslash\textbf{Ours} &\textbf{88.189}&\textbf{33.992}&\textbf{88.191}&\textbf{34.004}&\textbf{0.7890}&\textbf{0.1924}&\textbf{0.7891}&\textbf{0.1924}\\ 
\midrule
\multirow{5}{*}{\centering\arraybackslash\textbf{Chicago}}& NT-WRMSE &\underline{94.132}&\underline{19.261}&93.906&16.160&\textbf{0.8928}&0.0747&0.8661&0.0618\\ 
& AT-Random &94.071&18.426&93.954&16.816&0.8897&\underline{0.0890}&0.8803&0.0840\\ 
& AT-Degree &94.054&18.187&93.989&17.293&0.8895&0.0887&0.8817&\underline{0.0868}\\ 
& AT-TNDS & 94.028 &17.829&\underline{94.006}&\underline{17.531}&0.8898&0.0618&\underline{0.8854}&0.0566\\ 
& \centering\arraybackslash\textbf{Ours} &\textbf{94.231}&\textbf{20.632}&\textbf{94.231}&\textbf{20.632}&\underline{0.8908}&\textbf{0.0980}&\textbf{0.8908}&\textbf{0.0981}\\ 
\bottomrule
\end{tabular}}
\caption{Evaluation of the robustness of spatiotemporal graph adversarial training techniques based on GSNet. The table provides a detailed analysis of natural and robust performance, with robustness assessed against the STPGD-TNDS attack. The evaluation metrics include Rec-maj, Rec-min, MAP-maj, and MAP-min, with the best results highlighted in bold and the second-best results underlined.}
\label{tab:robustness}
\end{table*}

\section{Experiments}

\subsection{Datasets and Baselines}
To evaluate the effectiveness of our proposed MinGRE, we
conduct experiments on two benchmark datasets, including \textbf{NYC} and \textbf{Chicago}. The NYC and Chicago datasets contain finely-grained and sparse urban accident data, making them particularly well-suited for studying SGL models under ZID \cite{GSNet}. The detailed information on datasets is summarized in Table 1 of the Appendix.

We evaluated the adversarial robustness of our model by comparing it with various attack strategies: \textbf{STPGD-Random}, \textbf{STPGD-Degree}, \textbf{STPGD-Pagerank}, and the state-of-the-art \textbf{STPGD-TNDS} from Liu et al. \cite{TNDS}. Our method was benchmarked against spatiotemporal adversarial training methods: \textbf{AT-Random}, \textbf{AT-Degree}, \textbf{AT-Pagerank}, and \textbf{AT-TNDS} \cite{RDAT}. We also examined the effectiveness of different loss functions—\textbf{WRMSE} \cite{GSNet}, \textbf{NBL} \cite{NB}, and \textbf{BMSE} \cite{balancedmse}—on target models \textbf{GSNet} \cite{GSNet} and \textbf{Graph WaveNet} \cite{GWN}.

\subsection{Evaluations}
Building on \cite{GSNet}, we evaluate model performance from a ranking perspective by calculating recall and precision for majority and minority classes under ZID. We use \textbf{Rec-maj}, \textbf{Rec-min} to quantify the overlap between predicted and actual zero, non-zero observations. Ranking quality is further assessed using Mean Average Precision (MAP) for the top-k matches (\textbf{MAP-maj}, \textbf{MAP-min}). Performance disparity is represented by the difference between zero and non-zero observations (\textbf{Rec-D}, \textbf{MAP-D}). These metrics are commonly employed to gauge accuracy and robustness disparity \cite{accdisp,fairAT, NEURIPS2022RobFair}.

\begin{table*}[t]
\centering
\setlength{\tabcolsep}{4pt} 
\renewcommand{\arraystretch}{0.9} 
\resizebox{\textwidth}{!}{
\begin{tabular}{c c c c c c c c c c c c c}
\toprule
\multirow{2}{*}{\centering\arraybackslash\textbf{Dataset}} & \multicolumn{2}{c}{\textbf{Attacks}} & \multicolumn{2}{c}{\textbf{Natural}} & \multicolumn{2}{c}{\textbf{STPGD-Random}} & \multicolumn{2}{c}{\textbf{STPGD-Degree}} & \multicolumn{2}{c}{\textbf{STPGD-Pagerank}} & \multicolumn{2}{c}{\textbf{STPGD-TNDS}} \\ 
\cmidrule(lr){2-13}
  & \multicolumn{2}{c}{\textbf{Metrics}} & \textbf{Rec-D} & \textbf{MAP-D} & \textbf{Rec-D} & \textbf{MAP-D} & \textbf{Rec-D} & \textbf{MAP-D} & \textbf{Rec-D} & \textbf{MAP-D} & \textbf{Rec-D} & \textbf{MAP-D} \\ 
\midrule
\multirow{8}{*}{\centering\arraybackslash\textbf{NYC}} & \multirow{3}{*}{\centering\arraybackslash\textbf{ZID}} & NT-WRMSE &54.23&0.5978&59.92&0.6188&59.78&0.6184&59.82&0.6182 &59.60&0.6113\\ 
 &  & NT-NBL &\textbf{53.37}&\textbf{0.5933}&\textbf{52.30}&0.6115&\textbf{53.83}&0.6043&54.38&0.6124 &54.60&0.6136 \\ 
 &  & NT-BMSE &\underline{53.59}&0.5959&54.21&\underline{0.5987}&54.20&\underline{0.5995}&\underline{54.26}&\underline{0.5991}&\underline{54.25}&\underline{0.5989}\\ 
\cmidrule(lr){2-13}
 & \multirow{3}{*}{\centering\arraybackslash\textbf{STAT}} & AT-Random &55.58&0.5991&56.88&0.6040&57.34&0.6010&57.40&0.6034&57.16&0.6051\\ 
 &  & AT-Degree &55.72&0.5977&56.61&0.6033&56.93&0.6043&56.89&0.6029&56.89&0.6055\\ 
 &  & AT-TNDS &56.21&0.6105&56.43&0.6354&56.32&0.6358&56.37&0.6337&56.43&0.6229\\ 
\cmidrule(lr){2-13}
 & & \centering\arraybackslash\textbf{Ours} &54.20&\underline{0.5966}&\underline{54.19}&\textbf{0.5967}&\underline{54.19}&\textbf{0.5966}&\textbf{54.19}&\textbf{0.5966}&\textbf{54.19}&\textbf{0.5967}\\ 
\midrule
\multirow{8}{*}{\centering\arraybackslash\textbf{Chicago}} & \multirow{3}{*}{\centering\arraybackslash\textbf{ZID}} & NT-WRMSE &74.87&0.8181&77.86&0.8063&77.64&0.8060&77.75&0.8062 &77.75&0.8043\\ 
 &  & NT-NBL &74.43&0.7953&74.43&\underline{0.7953}&74.43&0.7953&74.43&0.7953 &74.43&0.7953\\ 
 &  & NT-BMSE &\textbf{72.77}&\textbf{0.7920}&\underline{73.65}&0.8056&\textbf{73.54}&0.8025&\textbf{73.54}&0.8021&\underline{73.71}&0.8095\\ 
\cmidrule(lr){2-13}
 & \multirow{3}{*}{\centering\arraybackslash\textbf{STAT}} & AT-Random &75.65&0.8007&76.97&0.7973&76.81&\underline{0.7938}&76.97&\underline{0.7945}&77.14&0.7963\\ 
 &  & AT-Degree &75.87&0.8008&76.59&0.7956&76.81&0.7947&76.70&0.7947&76.70&\underline{0.7949}\\ 
 &  & AT-TNDS &76.20& 0.8280&76.53&0.8289&76.36&0.8287&76.42&0.8305&76.47&0.8288\\ 
\cmidrule(lr){2-13}
 & & \centering\arraybackslash\textbf{Ours} &\underline{73.60}&\underline{0.7928}&\textbf{73.60}&\textbf{0.7927}&\underline{73.60}&\textbf{0.7927}&\underline{73.60}&\textbf{0.7928}&\textbf{73.60}&\textbf{0.7927}\\ 
\bottomrule
\end{tabular}}
\caption{Evaluation of the performance disparity of spatiotemporal graph adversarial training techniques based on GSNet. The table provides a detailed analysis of natural and robust performance disparity under various attacks. The evaluation metrics include Rec-D and MAP-D, with the best results highlighted in bold and the second-best results underlined.}
\label{tab:disparity}
\end{table*}

\subsection{Main Results}
We conduct a comprehensive analysis from three perspectives: robustness, performance disparity, and the effectiveness of sub-modules.
\subsubsection{Robustness Analysis}
Table \ref{tab:robustness} summarizes the natural and robust performance of various spatiotemporal adversarial training methods on the NYC and Chicago datasets. Key insights include: 1) Under the STPGD attack, the NT-WRMSE method shows significant declines in Rec-maj, Rec-min, MAP-maj, and MAP-min on the NYC dataset by approximately 1.3\%, 19.3\%, 3.4\%, and 21.5\%, respectively. This highlights the critical need to enhance SGL models' robustness across all classes under ZID scenarios. 2) Our method demonstrates superior robustness, particularly in minority classes, surpassing the state-of-the-art AT-TNDS by approximately 0.4\%, 9.8\%, 2.5\%, and 31.9\% in Rec-maj, Rec-min, MAP-maj, and MAP-min on the NYC dataset. While AT-TNDS achieves strong average robustness, it falls short in minority class protection, revealing the limitations of gradient-based victim selection strategies under zero-inflation contexts. The inherent gradient bias \cite{imbgra} leads to skewed adversarial examples, impeding uniform robustness enhancement.

\begin{figure}[htbp]
  \centering
  \includegraphics[width=\columnwidth]{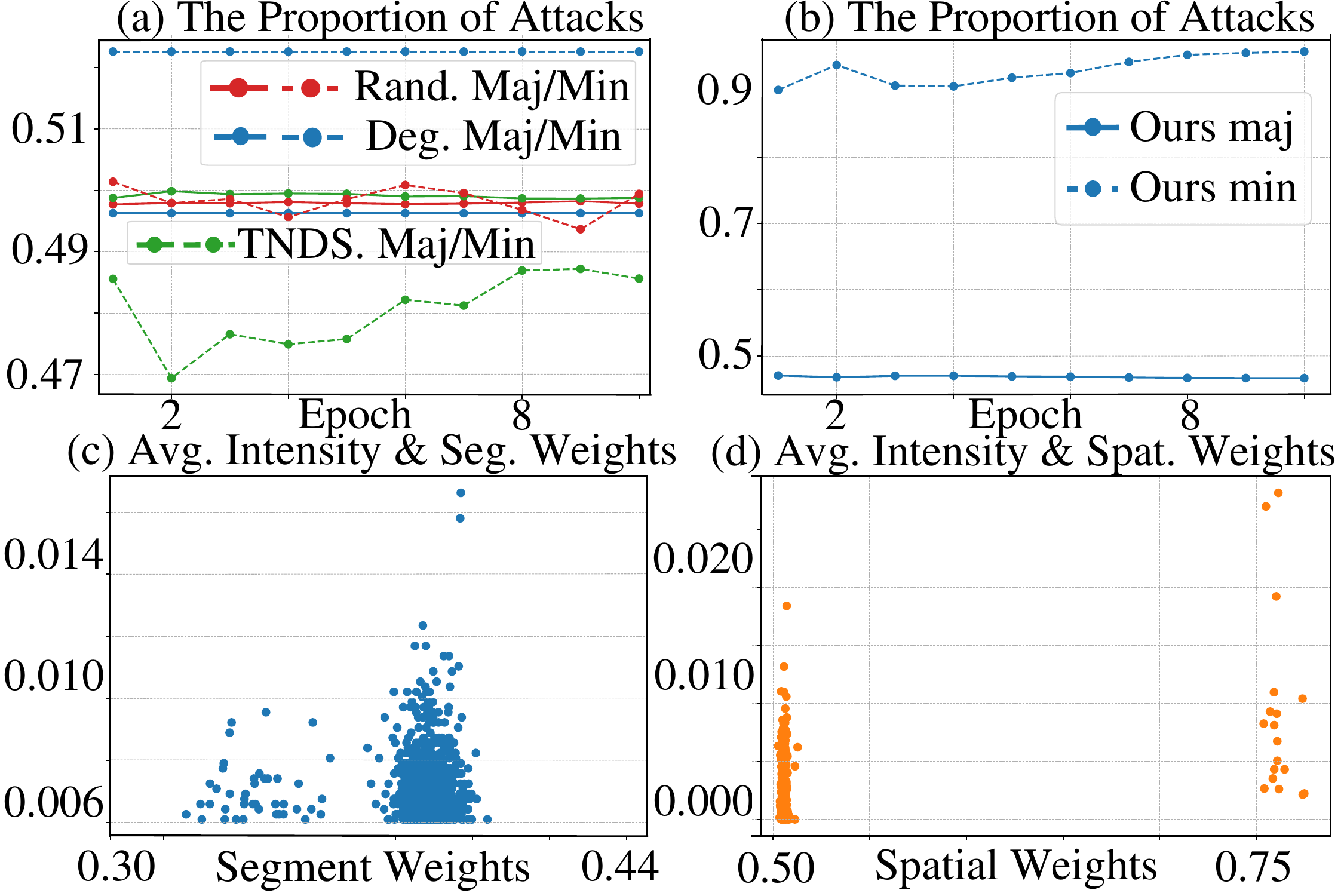}
  \caption{Effectiveness of sub-modules on NYC datasets.}
  \label{fig:results}
\end{figure}

\subsubsection{Peformance Disparity Analysis}
In Table 2, we evaluate the performance disparity of various spatiotemporal adversarial training methods and zero-inflation distribution approaches on the NYC and Chicago datasets, yielding three main conclusions: 1) Spatiotemporal adversarial training, while boosting robustness, often increases the performance disparity between majority and minority classes. For example, AT-TNDS raises Rec-D and MAP-D by 3.7\% and 2.1\% on the NYC dataset, mainly due to the decline in minority class performance, highlighting the need to address this issue. 2) ZID methods reduce natural performance disparities, as seen in the comparison of NT-NBL and NT-BMSE, though they do not consistently achieve optimal robust performance. On the NYC dataset, these methods sometimes underperform compared to adversarial training in terms of robust disparity. 3) Our method achieves the lowest natural and robust performance disparities, reducing Rec-D and MAP-D by 3.6\% and 2.3\% on the clean NYC dataset, and by 4.0\% and 4.2\% on the perturbed NYC dataset, compared to AT-TNDS.

\begin{figure}[htbp]
  \centering
  \includegraphics[width=\columnwidth]{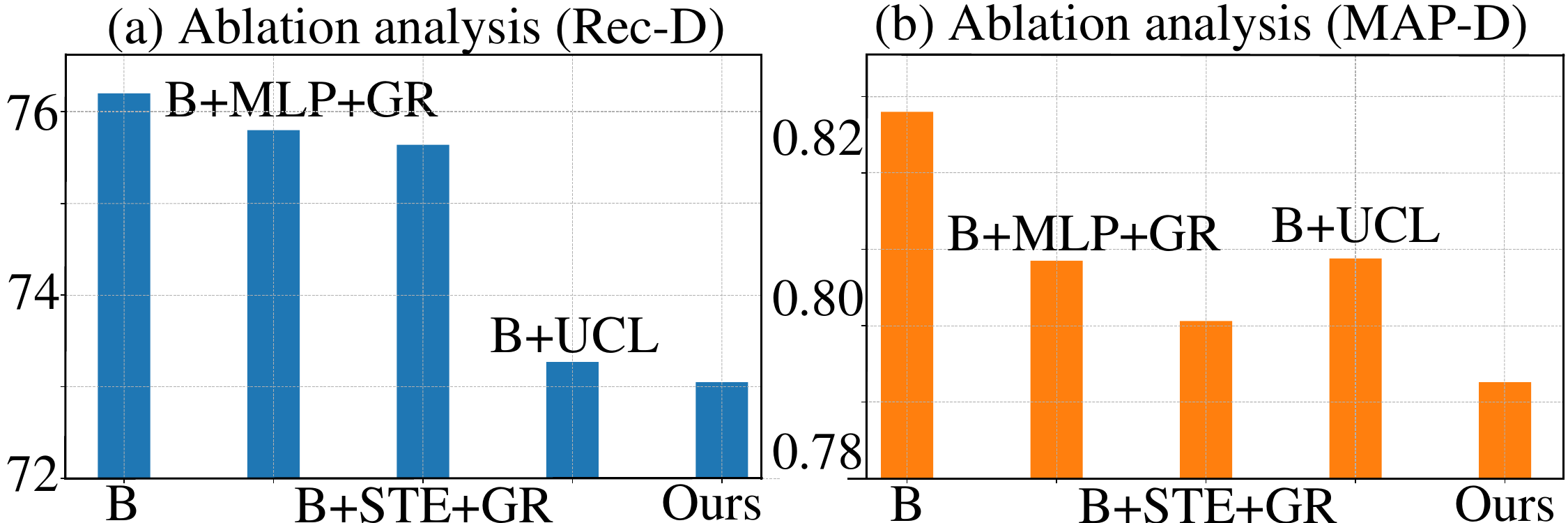}
  \caption{Ablation studies of the proposed Adversarial Examples Generation Module and Uncertainty-guided Contrastive Loss Module on Chicago datasets.}
  \label{fig:ablation}
\end{figure}

\subsubsection{Effectiveness of Sub-Modules}
To assess the efficacy of our proposed module, we initially visualized the adversarial sample generation process, noting a reduced gradient disparity between minority and majority classes (see Figure \ref{fig:emprical}). This recalibration introduces a greater number of minority samples into adversarial training,  contrasting with the AT-TNDS method that included the fewest (see Figure \ref{fig:results}). Furthermore, segment and spatial attention matrix analyses revealed that segments with frequent non-zero events of high intensity garnered higher weights (see Figure \ref{fig:results}), indicating our mechanism’s proficiency in capturing event frequency and intensity. Lastly, random visualizations of the feature space also showed improved separability (see Figure \ref{fig:emprical}), with a slight increase in minority class contour coefficients (from -0.006 to 0.1), reflecting the inherent difficulty in differentiating the data.

\subsection{Ablation Study}
We conduct ablation studies on the Chicago datasets to validate the proposed adversarial example generation and uncertainty-guided contrastive loss (UCL) modules. The baseline model (B) is a spatiotemporal adversarial training method (AT-TNDS) with weighted RMSE loss. STE, GR, and UCL represent the spatiotemporal encoder, gradient reweighting, and loss module, respectively. From Figure \ref{fig:ablation}, we observe three conclusions as follows. 1) Gradient reweighting reduces performance disparity by more effectively selecting minority instances, while the spatiotemporal encoder enhances performance through the capture of cross-segment dependencies. 2) The "B+UCL" variant enhances feature separability, outperforming other methods on Rec-D. 3) Integrating gradient reweighting and UCL achieves the lowest performance disparity, confirming the effectiveness of the proposed modules.

\section{Conclusion}
In summary, our study highlights the critical need to address performance disparities in spatiotemporal graph learning under zero-inflated distributions for urban risk management \cite{TKDEUrban}. We show that traditional adversarial training worsens the performance gap between majority and minority classes, while our MinGRE framework reduces this disparity and improves model robustness. Visualizations and ablation studies confirm MinGRE's effectiveness in recalibrating gradients, enhancing inter-class separability, and accurately capturing non-zero events. These results emphasize MinGRE's potential to advance more equitable and robust models \cite{AAAIRobFair, NeuralcomputingRobFair} for urban risk management.

\section{Acknowledgments}
This work was supported in part by the National Natural Science Foundation of China under Grant 72434005, Grant 72225011 and Grant 72293575.

\bibliography{aaai25}


\end{document}